\documentclass[letterpaper, 10 pt, conference]{ieeeconf}

\IEEEoverridecommandlockouts                    

\usepackage[labelformat=simple]{subcaption}

\usepackage[font=small]{caption}
\usepackage{graphicx}
\usepackage{tabularx}
\usepackage{times}
\usepackage{amsmath}
\usepackage{amssymb}
\usepackage{comment}
\usepackage{url}
\usepackage{multirow}
\usepackage{colortbl}
\usepackage{booktabs}
\usepackage[table,xcdraw]{xcolor}
\usepackage[normalem]{ulem}
\newcolumntype{x}[1]{>{\centering\arraybackslash}p{#1pt}}
\newcolumntype{y}[1]{>{\raggedright\arraybackslash}p{#1pt}}
\newcolumntype{z}[1]{>{\raggedleft\arraybackslash}p{#1pt}}
\usepackage{xcolor}
\usepackage{pifont}
\usepackage{array}
\usepackage{xspace}
\usepackage{amsmath}
\usepackage{nicematrix}

\usepackage{makecell}
\usepackage{threeparttable}
\usepackage{booktabs}
\usepackage{booktabs}
\usepackage{makecell}
\usepackage{threeparttable}
\usepackage{graphicx}
\usepackage{amssymb}

\makeatletter
\let\NAT@parse\undefined
\makeatother
\usepackage{hyperref}
\usepackage{booktabs}
\usepackage{threeparttable}

\useunder{\uline}{\ul}{}
\useunder{\cellcolor[HTML]{EFEFEF}}{\hl}{}

\newcommand{\nickname}{\textbf{\textit{FLUX}}\xspace}
\newcommand{\datasetname}{\textbf{\textit{DynBench}}\xspace}

\definecolor{best}{rgb}{1.0, 0.6, 0}
\definecolor{best2}{rgb}{1.0, 0.8, 0.6}

\title{\bf FLUX: Accelerating Cross-Embodiment Generative Navigation Policies via Rectified Flow and Static-to-Dynamic Learning
}

\author{
    Zeying Gong$^{*1}$~~\quad
    Yangyi Zhong$^{*1}$~~\quad
    Yiyi Ding$^{*1}$~~\quad 
    Tianshuai Hu$^2$~~\quad
    \\
    Guoyang Zhao$^1$~~\quad
    Lingdong Kong$^3$~~\quad
    Rong Li$^1$~~\quad
    Jiadi You$^1$~~\quad
    Junwei Liang$^{1,2,\dagger}$ 
    \thanks{$^*$ These authors contributed equally to this work..}
    \thanks{$^1$ The Hong Kong University of Science and Technology (Guangzhou)}
    \thanks{$^2$ The Hong Kong University of Science and Technology}
    \thanks{$^3$ National University of Singapore}
    \thanks{$^\dagger$ Corresponding author. Email: {\tt\footnotesize junweiliang@hkust-gz.edu.cn}}
}

\begin{document}

\maketitle
\thispagestyle{empty}
\pagestyle{empty}

\begin{abstract}
Autonomous navigation requires a broad spectrum of skills, from static goal-reaching to dynamic social traversal, yet evaluation remains fragmented across disparate protocols. We introduce \datasetname, a dynamic navigation benchmark featuring physically valid crowd simulation. Combined with existing static protocols, it supports comprehensive evaluation across six fundamental navigation tasks. Within this framework, we propose \nickname, the first flow-based unified navigation policy. By linearizing probability flow, \nickname replaces iterative denoising with straight-line trajectories, improving per-step inference efficiency by 47\% over prior flow-based methods and 29\% over diffusion-based ones. Following a static-to-dynamic curriculum, \nickname initially establishes geometric priors and is subsequently refined through reinforcement learning in dynamic social environments. This regime not only strengthens socially-aware navigation but also enhances static task robustness by capturing recovery behaviors through stochastic action distributions. \nickname achieves state-of-the-art performance across all tasks and demonstrates zero-shot sim-to-real transfer on wheeled, quadrupedal, and humanoid platforms without any fine-tuning. The project website is at \href{https://zeying-gong.github.io/projects/flux/}{\textcolor{blue}{\url{https://zeying-gong.github.io/projects/flux/}}}.
\end{abstract}

\section{Introduction}
Real-world visual navigation requires a broad spectrum of skills, from static goal-reaching and exploration to safe traversal in dynamic, human-populated spaces. These competencies are essential for reliable autonomy and higher-level planning. However, research remains deeply fragmented: static-environment policies are seldom tested under social constraints, while social navigation methods lack evaluation across diverse task modalities and heterogeneous platforms. This lack of a unified protocol not only hinders systematic benchmarking but also obscures the underlying relationship between these seemingly disparate navigation skills.

Existing approaches reflect this fragmentation in their design. Traditional reinforcement learning (RL)~\cite{wijmans2019dd, gong2025cognition} achieves strong task-specific performance but lacks the flexibility to generalize across different task diversities. Conversely, imitation learning (IL)~\cite{shah2022gnm, shah2023vint} demonstrates cross-task generalization yet remains ill-equipped for dynamic social scenarios. Static training distributions provide insufficient coverage of the recovery behaviors required for pedestrian interaction. Recently, generative modeling~\cite{sridhar2024nomad,gode2024flownav,cai2025navdp} has emerged as a promising paradigm to unify these capabilities by capturing multi-modal response spaces. However, their practical utility is limited by high inference latency, and the fundamental question of how training distributions shape cross-task generalization remains unexplored.

\begin{figure}[t]
    \centering
    \includegraphics[width=\linewidth]{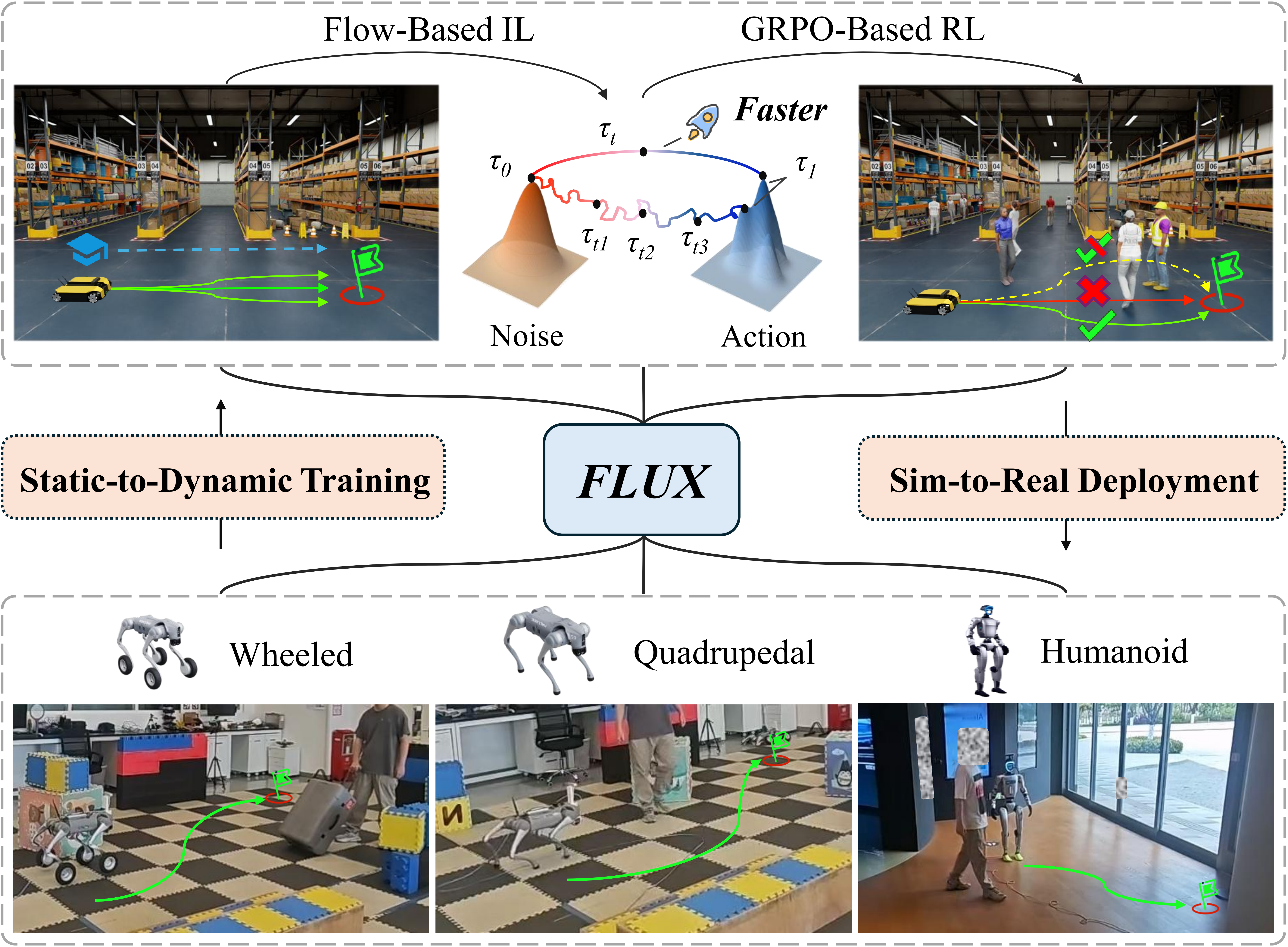}
    \vspace{-0.4cm}
    \caption{\textbf{\nickname: A Flow-Based Unified Policy for Cross-Embodiment Navigation.} Our static-to-dynamic training curriculum enables efficient, socially-aware navigation, which transfers zero-shot across three heterogeneous platforms in the real world.}
    \vspace{-0.4cm}
    \label{fig:teaser}
\end{figure}


A key insight motivates our approach to addressing these limitations: static environments are a degenerate special case of dynamic ones. We posit that a policy initially grounded in geometric priors via static imitation learning can achieve greater robustness when subsequently exposed to the high-dimensional challenges of human interaction. To operationalize this progressive perspective, we introduce \datasetname, a benchmark featuring physically-valid crowd simulation that complements existing static protocols. This unified framework enables a two-stage training paradigm where navigation in dynamic environments reinforces predictive avoidance strategies that are overlooked in static datasets.

Therefore, we propose \nickname, the first \underline{FL}ow-based \underline{U}nified policy for \underline{X}-platform (Cross-Embodiment) navigation, as shown in Fig.~\ref{fig:teaser}. By linearizing probability flow, \nickname replaces the iterative denoising of diffusion models with straight-line trajectories, reducing sampling steps while achieving superior performance over diffusion baselines. Initially established through static IL, \nickname undergoes refinement using Group Relative Policy Optimization (GRPO)~\cite{shao2024deepseekmath} within \datasetname's dynamic environments. Through this transition from static to dynamic, \nickname not only masters socially-aware navigation, but yields a positive transfer effect that enhances performance on static tasks. We validate \nickname through extensive simulations, achieving state-of-the-art (SOTA) performance across all tasks and sim-to-real generalizability across wheeled (Unitree Go2-W), quadrupedal (Unitree Go2), and humanoid (Unitree G1) robots.

Overall, our main contributions are as follows:
\begin{itemize}
    \item 
    We propose \nickname, the first flow-based unified navigation policy that achieves \textbf{47\%} higher efficiency than prior flow methods, bridging the gap between multi-modal trajectory generation and efficient robotic control.
    
    \item
    We introduce \datasetname, a benchmark with physically-valid crowd simulation that enables a two-stage IL-then-RL training paradigm, revealing that dynamic training yields a positive transfer effect on static tasks.
    
    \item
    \nickname achieves state-of-the-art performance over all six tasks in simulation, and demonstrates zero-shot sim-to-real transfer across wheeled, quadrupedal, and humanoid platforms without any real-world fine-tuning.
\end{itemize}

\section{Related Work}

\subsection{Deterministic Navigation Policies.}
Traditional visual navigation largely relies on deterministic paradigms mapping observations to single actions. While classical planners~\cite{fox2002dynamic, yamauchi1997frontier} offer interpretability, they are often hindered by numerous handcrafted parameters. End-to-end RL~\cite{wijmans2019dd, gong2025cognition} and hybrid modular approaches~\cite{yang2023iplanner, roth2024viplanner} achieve task-specific success but struggle with cross-task generalization or social constraints, while recent IL approaches~\cite{shah2022gnm, shah2023vint} remain primarily confined to static scenes. Crucially, these policies collapse complex action distributions into single-point estimates, failing to capture the multi-modality required for safe recovery and proactive avoidance in crowded spaces. This motivates us to adopt a generative paradigm and propose \nickname.

\subsection{Generative Navigation Models.}
Generative policies~\cite{ho2020denoising} provide a robust alternative to deterministic mapping by learning multi-modal action distributions. While diffusion-based models like NoMaD~\cite{sridhar2024nomad} and NavDP~\cite{cai2025navdp} excel at capturing diverse trajectories, their iterative denoising limits real-time deployment. To mitigate this, flow matching~\cite{lipman2022flow, liu2022flow} has emerged; for instance, FlowNav~\cite{gode2024flownav} achieves faster sampling but remains restricted to static environments and lacks support for coordinate-based goals. Furthermore, the fundamental necessity of stochasticity for navigation robustness remains under-explored. \nickname bridges these gaps by linearizing probability flow for efficient inference, while our experiments demonstrate that such generative modeling is essential for robust navigation.

\subsection{Embodied Navigation Benchmarks.}
Beyond algorithmic design, evaluating general navigation capabilities is hindered by fragmented benchmarks. Photorealistic datasets like Gibson~\cite{xia2018gibson} provide static environments but exclude social interaction. Conversely, social navigation suites~\cite{chen2019crowd, kastner2022arena, gong2025cognition} incorporate pedestrians but often rely on oversimplified motion models or narrow task modalities. While real-world datasets like ~\cite{karnan2022socially} offer high fidelity, it lacks the controllability required for reproducible benchmarking. \datasetname bridges this divide by integrating physically-valid crowd simulation with established static protocols~\cite{cai2025navdp}, providing a unified framework to evaluate general navigation across diverse tasks and platforms.

\begin{figure*}[t]
    \centering
    \includegraphics[width=\linewidth]{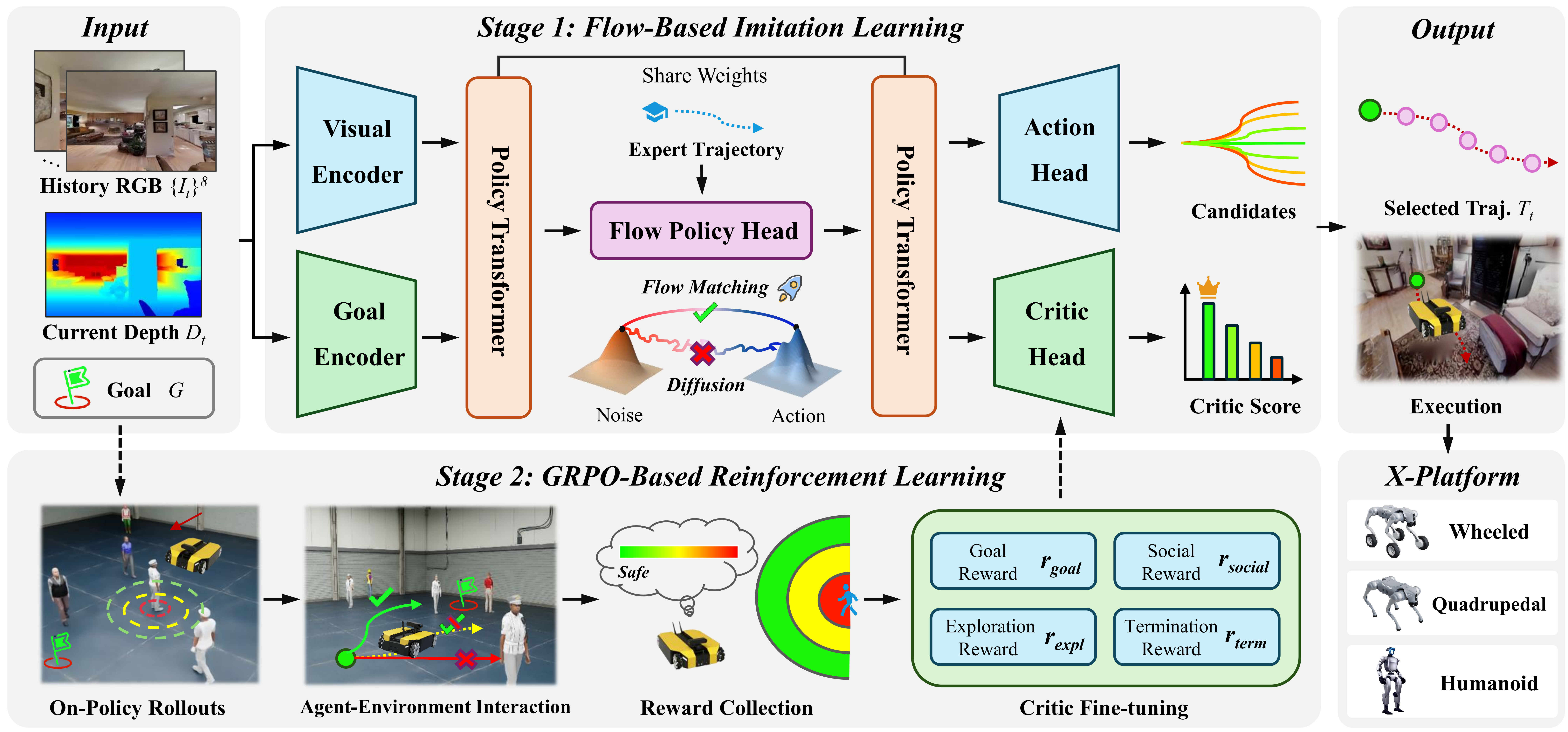}
    \vspace{-0.4cm}
    \caption{\textbf{Overview of \nickname framework.} \nickname follows a static-to-dynamic training paradigm. Stage 1 (Top): Given egocentric visual observations and a goal, the flow policy head is pre-trained via imitation learning on static expert trajectories. It generates diverse candidate paths, which are evaluated by the critic head. Stage 2 (Bottom): The framework is post-training using Group Relative Policy Optimization via on-policy rollouts. This stage optimizes for both goal-reaching efficiency and social compliance in dynamic environments.}
    \label{fig:main}
    \vspace{-0.4cm}
\end{figure*}

\section{PROBLEM FORMULATION}
\label{chapter:problem}

We consider a robot navigating among $N \geq 0$ dynamic agents, seeking an optimal trajectory $\tau^*$ within the collision-free trajectory set $\mathcal{T}$. We focus on two main task categories.
 
\noindent\textbf{Goal-conditioned Navigation.} 
The robot aims to reach a target $g \in \mathcal{G}$ defined by the specific task modality. $\mathcal{G}$ encompasses static objectives, including Point Navigation (\textit{PointNav}) and Image Navigation (\textit{ImageNav}), as well as Dynamic Point Navigation (\textit{Dyn. PointNav}), where $g(t)$ corresponds to the real-time coordinates of a moving agent. The objective is to find an optimal trajectory $\tau^*$:
\begin{equation}
\tau^* = \arg\min_{\tau \in \mathcal{T}} \; c_{goal}(\tau, g) + \beta \cdot c_{social}(\tau, \tau_{1:N})~,
\end{equation}
where $c_{goal}$ denotes goal-reaching efficiency, $c_{social}$ penalizes violations of proxemic norms relative to dynamic agents, and $\beta \geq 0$ is a weighting factor. This formulation includes \textit{PointNav}, \textit{ImageNav}, and \textit{Dyn. PointNav} as special cases ($\beta{=}0$) and Social Navigation (\textit{SocialNav}) ($\beta{>}0$ and $N{>}0$).

\noindent\textbf{Coverage Maximization.} 
In exploration, the robot maximizes the explored area $\mathcal{E}(\tau)$ without an explicit goal:
\begin{equation}
 \tau^* = \arg\max_{\tau \in \mathcal{T}} \mathcal{E}(\tau)~.
\end{equation}
This formulation unifies \textit{Exploration} ($N{=}0$) and Dynamic Exploration (\textit{Dyn. Exploration}) ($N{>}0$).


\section{METHODOLOGY}
\label{chapter:method}
\subsection{Overview}

\nickname is a generative navigation policy that maps ego-centric observations to a diverse set of trajectory candidates. We formulate the policy as a conditional generative model that transforms a latent noise $\tau_0 \sim \mathcal{N}(0, I)$ into a plausible future trajectory $\tau_1$. As shown in Fig.~\ref{fig:main}, the framework consists of two coupled stages: (1) \textbf{Flow-Based Imitation Learning} stage, which learns a linearized probability flow to enable high-quality trajectory sampling with significantly fewer steps than diffusion-based counterparts; and (2) \textbf{GRPO-Based Reinforcement Learning} stage, which fine-tunes the generative prior using the GRPO algorithm to enhance social safety and robustness in dynamic environments.

\subsection{Stage 1: Flow-Based Imitation Learning}
Prior diffusion-based planners~\cite{sridhar2024nomad,cai2025navdp} adopt DDPM with $\epsilon$-prediction, where the network $\epsilon_\theta$ predicts the injected noise at each denoising step $k$:
\begin{equation}
\tau_{k-1} = \frac{1}{\sqrt{\alpha_k}}\!\left(\tau_k - 
\frac{1-\alpha_k}{\sqrt{1-\bar{\alpha}_k}}
\epsilon_\theta(\tau_k, k, \mathbf{e}_t)\right) + \sigma_k z~,
\end{equation}
where $\alpha_k, \bar{\alpha}_k$ are noise schedule coefficients, $\sigma_k z$ 
($z\!\sim\!\mathcal{N}(0,I)$) is per-step stochastic noise, and $\mathbf{e}_t = f_{\text{enc}}(\mathcal{X}_t)$ is the conditional embedding from observation context $\mathcal{X}_t$. This process has two drawbacks: (1) per-step 
noise $\sigma_k z$ induces a curved stochastic transport path, increasing fitting difficulty; (2) $\epsilon$-prediction provides only indirect supervision of the target trajectory. While a deterministic $x_0$-regression $\hat{\tau} = f_{\text{reg}}(\mathbf{e}_t)$ avoids these issues~\cite{he2024lotus}, it lacks the multi-modal diversity that is critical for deadlock recovery, leading to inferior performance (see Tab.~\ref{tab:ablation}).

These motivate us to adopt flow-based generation. Conditional Flow Matching (CFM)~\cite{lipman2022flow} addresses the curvature issue by learning a velocity field $v_\theta$ that transports noise to data along a probability path. Given noise $\tau_0$ and expert trajectory $\tau_1$, CFM constructs an interpolant $\tau_t$ and minimizes:
\begin{equation}
\mathcal{L}_{\text{CFM}} = \mathbb{E}_{t,\tau_0,\tau_1}
\!\left[\|v_\theta(t, \tau_t, \mathbf{e}_t) - 
u_t(\tau_t \mid \tau_1)\|_2^2\right]~,
\end{equation}
where $u_t(\tau_t \mid \tau_1)$ is the conditional target velocity defined by the chosen interpolant. CFM already yields competitive navigation performance, but places no constraint on the transport path, leaving residual curvature that still demands a larger number of function evaluations (NFE) at inference. We therefore adopt Rectified Flow (RF)~\cite{liu2022flow}, which instantiates CFM with a linear interpolant:
\begin{equation}
\tau_t = (1-t)\tau_0 + t\tau_1~, \quad 
u_t = \tau_1 - \tau_0~,
\end{equation}
where $t\!\sim\!\mathcal{U}[0,1]$ is the flow time. The constant target velocity 
$u_t = \tau_1 - \tau_0$ enforces straight-line transport by construction, simplifying 
the learning target and yielding better generation quality under the same NFE budget. The network $v_\theta$ is optimized via:
\begin{equation}
\mathcal{L}_{\text{RF}} = \mathbb{E}_{t,\tau_0,\tau_1}
\!\left[\|v_\theta(t, \tau_t, \mathbf{e}_t) - 
(\tau_1 - \tau_0)\|_2^2\right].
\end{equation}
At inference, we integrate the learned velocity field with $K$ Euler steps to 
generate each trajectory candidate:
\begin{equation}
\tau^{(k+1)} = \tau^{(k)} + \tfrac{1}{K}\,v_\theta\!\left(
\tfrac{k}{K},\,\tau^{(k)},\,\mathbf{e}_t\right),~~ k=0,\dots,K{-}1 ~,
\end{equation}
where $\tau^{(0)} = \tau_0$ and $\tau^{(K)}$ denotes the final predicted trajectory. In total, $M$ candidates are generated and evaluated via a shared critic head to select the optimal trajectory. The effects of the candidate number $M$ and the number of sampling steps $K$ are analyzed in Tab.~\ref{tab:efficiency}.

\subsection{Stage 2: GRPO-Based Reinforcement Learning}
IL on static datasets prioritizes path efficiency over social compliance, leading to discomfort or collision risks in human-shared environments. We address this by fine-tuning \nickname~via GRPO across three tasks: \textit{SocialNav}, 
\textit{Dyn.~PointNav}, and \textit{Dyn.~Exploration}. Crucially, the $M$ candidates produced by our generative backbone serve as a natural reference group for group-relative advantage estimation~\cite{shao2024deepseekmath}, eliminating the need for a separate value-function critic and enabling straightforward inference.

\begin{figure*}[t]
    \centering
    \includegraphics[width=\linewidth]{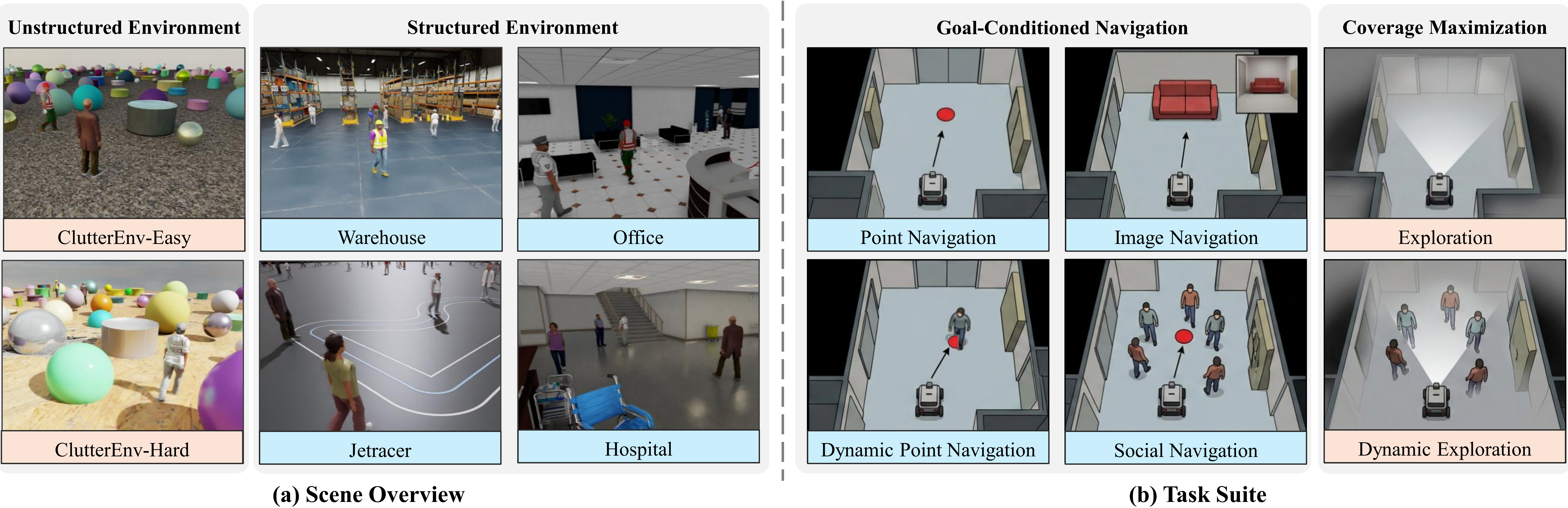}
    \vspace{-0.4cm}
    \caption{\textbf{Overview of \datasetname evaluation suite.} Our benchmark provides a unified framework for fundamental navigation. (a) Diverse environments spanning unstructured cluttered environments with randomized obstacles and structured Isaac Sim scenes. (b) Six task modalities categorized into goal-conditioned navigation and coverage maximization, supporting our static-to-dynamic learning paradigm.}
    \vspace{-0.4cm}
    \label{fig:benchmark}
\end{figure*}

\noindent\textbf{Reward Design.}
For goal-conditioned tasks (\textit{SocialNav} and \textit{Dyn.~PointNav}), 
the reward combines asymmetric goal progress with proxemic compliance:
\begin{equation}
r = r_{\text{goal}} + \alpha(\Delta d) \cdot (d_{t-1} - d_t) 
    + r_{\text{social}} + r_{\text{term}}~,
\end{equation}
where $d_t$ is the distance to goal at step $t$, $r_{\text{goal}}{=}20$ is the success bonus, and $\alpha(\Delta d)$ is an asymmetric progress weight ($\alpha{=}5$ when approaching, $\alpha{=}2$ when receding) to reduce oscillation while preserving goal attraction. $r_{\text{term}}$ applies $-2$ for getting stuck and $-0.5 \cdot d_t$ as a timeout penalty. The social penalty $r_{\text{social}}$ follows Hall's proxemic zones~\cite{hall1963system}:
\begin{equation}
r_{\text{social}} = \sum_{j=1}^{N}\begin{cases}
-0.5~, & d_j < d_{\text{int}}~, \\
-0.1 \cdot \dfrac{d_{\text{com}}-d_j}{d_{\text{com}}-d_{\text{int}}}~, 
    & d_{\text{int}} \leq d_j < d_{\text{com}}~, \\
0~, & \text{otherwise},
\end{cases}
\end{equation}
where $d_j$ is the distance to the $j$-th pedestrian, $d_{\text{int}}{=}0.45$\,m and $d_{\text{com}}{=}1.2$\,m are the intimate and comfort zone boundaries. The coefficients are chosen so that the worst-case social cost does not exceed the typical progress reward, preventing social avoidance from dominating navigation.

For \textit{Dyn.~Exploration}, we adopt a grid-based coverage reward rather than a direction-diversity penalty, which would otherwise incentivize non-directional oscillation:
\begin{equation}
r = r_{\text{expl}} + \lambda \cdot \phi(x_t) 
    + r_{\text{social}} + r_{\text{term}}~,
\end{equation}
where $r_{\text{expl}} = 4\|\Delta \mathbf{x}\|$ encourages moving and 
$r_{\text{term}}{=}-0.05$ discourages early termination. The environment is 
discretized into a $0.5$\,m grid via a per-episode hash map $g(\cdot)$. The novelty bonus $\phi(x_t)$ is $1.0$ for the first visit to a cell, $0.1$ for the second, and $0$ thereafter. We set $\lambda{=}2$ such that discovering a new cell yields a reward comparable to $0.25$\,m of forward motion, offering a moderate signal of curiosity.

\noindent\textbf{Episode-level Advantage Estimation} Step-based policy gradient updates suffer from credit assignment failures: when experience buffers are truncated at fixed intervals, the cumulative effect of early actions, such as whether the agent ultimately reaches the goal, cannot be fully captured. To address this, we trigger gradient updates only at episode boundaries. Specifically, we compute the discounted return $G_t$ over the whole episode:
\begin{equation}
    G_t = \sum_{k=t}^{T} \gamma^{k-t} \, r_k~,
\end{equation}
where $r_k$ is the reward at step $k$, $\gamma = 0.99$ is the discount factor, and $T$ is the episode length. The advantage $\hat{G}_t$ is then obtained by Z-score normalizing all per-step returns $\{G_t\}_{t=0}^{T}$ within the current episode, with clipping to $[-3, 3]$ to avoid unstable gradients from outliers.

Unlike original GRPO, which normalizes advantages across sampled responses to a single prompt, ours normalizes advantages per episode, adapting GRPO to sequential robot navigation. The resulting $\hat{G}_t$ serves directly as the advantage in the clipped policy gradient objective:
\begin{equation}
\mathcal{L}_{\mathrm{GRPO}} = -\mathbb{E}_{t} \!\left[\min\!\left(
            \rho_t \, \hat{G}_t,\;
            \mathrm{clip}\!\left(\rho_t,\, 1 - \varepsilon,\, 1 + \varepsilon\right) \hat{G}_t
        \right)
    \right]~,
\end{equation}
where $\rho_t$ is the importance sampling ratio and $\varepsilon$ is the clipping threshold. Through this multi-task RL fine-tuning, \nickname corrects the proximity-tolerance bias inherent in static datasets, where agents learn neither to anticipate moving obstacles nor to maintain safe margins. 

\section{Benchmark}
\label{chapter:benchmark}

To enable systematic evaluation of navigation policies across both static 
and dynamic regimes, we introduce \datasetname, a benchmark that extends 
existing static protocols~\cite{cai2025navdp} with human-populated scenarios, 
as shown in Fig.~\ref{fig:benchmark}. As summarized in Tab.~\ref{tab:benchmark_comparison}, existing benchmarks either adopt oversimplified pedestrian motion models (e.g., random walks or rule-based scripts) or rely on ORCA planners~\cite{van2011reciprocalorca} that assume perfect reciprocal cooperation, producing unnaturally 
coordinated crowd behaviors. \datasetname addresses these limitations through a hierarchical Markov Finite State Machine (FSM) behavior protocol combined with NavMesh-based global planning and geometry-based local collision avoidance, yielding more naturalistic pedestrian interactions within high-fidelity Isaac Sim~\cite{makoviychuk2021isaac} simulation.

\subsection{Scene Selection and Task Suite} 

Our benchmark comprises 20 unstructured scenes from NavDP's ClutterEnv for training (20,000 episodes) and 6 high-fidelity structured scenes for evaluation (600 episodes), built within NVIDIA Isaac Sim~\cite{makoviychuk2021isaac}. While the evaluation set is selective, it is designed to meet two critical 
technical requirements: (1) Large-scale architectural layouts capable of accommodating high-density crowds, and (2) Physically valid pedestrian dynamics enforced by the simulator's rigid-body physics engine. Specifically, pedestrian navigation combines NavMesh-based global path planning with geometry-based local collision avoidance, producing realistic and interactive crowd behaviors without relying on simplified reciprocal velocity assumptions (e.g., ORCA~\cite{van2011reciprocalorca}), which often yield unnaturally coordinated crowd motion.

Building upon the objectives defined in Sec.~\ref{chapter:problem}, we evaluate the following six tasks:
\begin{itemize}
\item \textbf{Static Tasks}: We adopt the standard \textit{PointNav}, \textit{ImageNav}, and \textit{Exploration} protocols from NavDP~\cite{cai2025navdp} to ensure fair comparisons.
\item \textbf{Dynamic Tasks}: We instantiate dynamic challenges:
\begin{itemize}
\item \textit{Dyn. PointNav}: The robot must track and reach the vicinity of a specific moving agent.
\item \textit{SocialNav}: The robot navigates to a static point goal within a high-density crowd (10--15 agents) while adhering to proxemic norms.
\item \textit{Dyn. Exploration}: The robot maximizes area coverage under dynamic disturbances.
\end{itemize}
\end{itemize}

\subsection{Structured Human Behavior Protocol} 

\begin{table}[t]
\centering
\caption{\textbf{Comparison of simulation benchmarks with dynamic humans.} \datasetname combines hierarchical Markov behavior modeling with NavMesh-based planning and geometry-based local avoidance, producing more naturalistic pedestrian interactions.}
\label{tab:benchmark_comparison}
\resizebox{\linewidth}{!}{
\begin{tabular}{l|ccccc}
\toprule
\textbf{Benchmark} & 
\textbf{Rendering} & 
\textbf{Physics} & 
\textbf{Behaviors} &
\textbf{Planners} \\
\midrule
CrowdNav~\cite{chen2019crowd}             & 2D & $\times$     & Rule-based  & ORCA~\cite{van2011reciprocalorca} \\
Arena-Bench~\cite{kastner2022arena}       & 2D\&3D & $\times$     & Random Walk & ORCA~\cite{van2011reciprocalorca} \\
Social-HM3D~\cite{gong2025cognition}      & 3D & $\times$     & Rule-based  & A*~\cite{hart1968formalastar} + ORCA~\cite{van2011reciprocalorca} \\
URBAN-SIM~\cite{wu2025towards}            & 3D & $\checkmark$ & Rule-based  & A*~\cite{hart1968formalastar} + ORCA~\cite{van2011reciprocalorca} \\
\midrule
\textbf{\datasetname{} (Ours)} & \textbf{3D} & \textbf{$\checkmark$} & \textbf{Markov FSM} & \textbf{NavMesh-A* + Geometric}~\cite{makoviychuk2021isaac} \\
\bottomrule
\end{tabular}
}
\vspace{-0.4cm}
\end{table}


To overcome the limitations of aimless wandering or predictable linear motion in existing benchmarks, we introduce a hierarchical behavior protocol.

\noindent\textbf{Global Trajectory Coherence.} To ensure long-horizon consistency, each pedestrian continuously traverses a fixed triangular path among three spatially distributed waypoints. Unlike the memoryless 
random walks in Arena~\cite{kastner2022arena} or simplistic linear oscillations in URBAN-SIM~\cite{wu2025towards}, our approach introduces multi-directional encounter geometries. By forcing pedestrians to change heading at each vertex, we require the robot to anticipate non-linear motion patterns and manage complex crossing angles. 

\noindent\textbf{Stochastic Action Transitions.} Unlike benchmarks where agents become static upon reaching a goal~\cite{gong2025cognition}, our pedestrians maintain persistent activity through three states: \textit{GoTo} (navigation), \textit{Idle} (pause), and \textit{LookAround} (rotation). Agents transition via a Markov matrix $P$:
\begin{equation}
P =
\begin{pNiceMatrix}[first-row, last-col=4]
\text{GoTo} & \text{Idle} & \text{LookAround} & \\
0 & 0.5 & 0.5 & \text{GoTo} \\
0.6 & 0 & 0.4 & \text{Idle} \\
0.7 & 0.3 & 0 & \text{LookAround}
\end{pNiceMatrix} .
\end{equation}

Moreover, a compounding challenge is velocity asymmetry: the robot is capped at $0.5$\,m/s to align with static navigation benchmark conventions and ensure deployment safety, while pedestrians move at $1.1$\,m/s consistent with real-world norms. This speed differential necessitates proactive spatial reasoning rather than reactive maneuvers.

\subsection{Evaluation Metrics and Criteria}
Beyond standard Success Rate (SR), Success weighted by Path Length (SPL), Exploration Time (ET), and Exploration Area (EA), we introduce a comprehensive evaluation framework for fundamental navigation, while the social metrics are grounded in Hall's Proxemics~\cite{hall1963system}:

\begin{itemize}
\item \textbf{Successful Trajectory Length (S-TL)}: Average path length of successful episodes, measuring relative efficiency in dynamic scenes where the shortest path is often unsafe and oracle path is hard to obtain.
\item \textbf{Social Compliance (SC)}: Percentage of time the agent stays inside the Social Space ($d < 1.2$\,m).
\item \textbf{Collision Count (Coll.)}: Average number of intimate space intrusions ($d < 0.45$\,m) per episode, serving as a proxy for physical collision risk.
\item \textbf{Minimum Distance (MinDist)}: The average of the minimum robot-to-human distances of within a full episode, quantifying avoidance tightness.
\end{itemize}

SR/SPL/SC are in \%; Coll. is a count per episode; ET in s; S-TL/MinDist in m; EA in m$^2$.

For both static and dynamic tasks, episodes terminate if the robot: 
(1) is \textbf{Stuck} (displaces less than $0.1$\,m for $>2.0$\,s); 
(2) \textbf{Timeout} ($>120$\,s); or for goal-conditioned tasks only: 
(3) \textbf{Success} (reaches within $1.0$\,m of the target).

\section{Experiments}
\label{chapter:Experiments}


\begin{table*}[ht]
    \centering
    \caption{\textbf{Performance comparison across six fundamental navigation tasks.} Our method achieves state-of-the-art performance. }
    \label{tab:nav_comparison}
    \resizebox{\linewidth}{!}{
    \begin{tabular}{lcc cc cc  cc cc cccc}
    \toprule
    \multirow{4}{*}{\textbf{Methods}} & \multicolumn{6}{c}{{\includegraphics[width=0.021\linewidth]{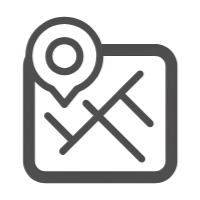}} \textbf{Static Scenes}} & \multicolumn{8}{c}{{\includegraphics[width=0.02\linewidth]{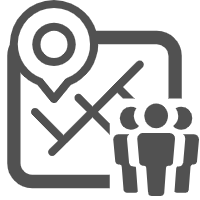}} \textbf{Dynamic Scenes}} 
    \\
    \cmidrule(lr){2-7} \cmidrule(lr){8-15} & \multicolumn{2}{c}{PointNav} & \multicolumn{2}{c}{Exploration} & \multicolumn{2}{c}{ImageNav} & \multicolumn{2}{c}{Dyn. PointNav} & \multicolumn{2}{c}{Dyn. Exploration} & \multicolumn{4}{c}{SocialNav} 
    \\
    \cmidrule(lr){2-3} \cmidrule(lr){4-5} \cmidrule(lr){6-7} \cmidrule(lr){8-9} \cmidrule(lr){10-11} \cmidrule(lr){12-15} & SR$\uparrow$ & SPL$\uparrow$ & ET$\uparrow$ & EA$\uparrow$ & SR$\uparrow$ & SPL$\uparrow$ & SR$\uparrow$ & S-TL$\downarrow$ & ET$\uparrow$ & EA$\uparrow$ & SR$\uparrow$ & Coll.$\downarrow$ & SC$\downarrow$ & MinDist.$\uparrow$ 
    \\
    \midrule
    \midrule
    \rowcolor{gray!10}\multicolumn{15}{l}{{\textcolor{gray}{\textbf{Traditional Planning Methods}}}} 
    \\
    $\bullet$~DWA~\cite{fox2002dynamic} & 33.5 & 32.4 & - & - & - & - &  38.2 & 18.5 & - & - & 41.8 & 10 & 7.3 & 4.1 
    \\
    $\bullet$~FBE~\cite{yamauchi1997frontier}& - & - & 4.3 & 11.2 & - & - & - & - &  20.3 & 39.5 & - & - & - & - 
    \\
    \midrule
    \rowcolor{gray!10}\multicolumn{15}{l}{{\textcolor{gray}{\textbf{Reinforcement Learning Methods}}}} 
    \\
    $\bullet$~DD-PPO~\cite{wijmans2019dd} & 19.5 & 19.2 & - & - & - & - & 8.4 & 7.7 & - & - & 1.5  & 18 & 4.3 & 3.4 
    \\
    $\bullet$~Falcon~\cite{gong2025cognition}& 40.0 & 33.6 & - & - & - & - & 19.0 & 16.4 & - & - & 13.8 & 17 & 4.2 & 3.6 
    \\
    \midrule
    \rowcolor{gray!10}\multicolumn{15}{l}{{\textcolor{gray}{\textbf{Hybrid Modular Methods}}}} 
    \\
    $\bullet$~iPlanner~\cite{yang2023iplanner}  & 66.8 & 65.1 & - & - & - & - & 30.8 & 15.4 & - & - & 35.3 & 13 & 6.3 & 4.3 
    \\
    $\bullet$~ViPlanner~\cite{roth2024viplanner} & 63.4 & 62.4 & - & - & - & - & 27.5 & 17.2 & - & - & 37.0 & 12 & 5.9 & 4.3 
    \\
    \midrule
    \rowcolor{gray!10}\multicolumn{15}{l}{{\textcolor{gray}{\textbf{Imitation Learning Methods}}}}  
    \\
    $\bullet$~GNM~\cite{shah2022gnm} & - & - & 22.1 & 27.8 & 16.3 & 15.7 & - & - &  29.7 & 65.5 & - & - & - & - 
    \\
    $\bullet$~ViNT~\cite{shah2023vint}  & - & - & 24.4 & 36.2 & 12.6 & 11.8 & - & - & 31.9 & 73.2 & - & - & - & - 
    \\
    \midrule
    \rowcolor{gray!10}\multicolumn{15}{l}{{\textcolor{gray}{\textbf{Generative Modeling Methods}}}} 
    \\
    $\bullet$~NoMaD~\cite{sridhar2024nomad} & - & - & 35.5 & 62.5 & 8.5 & 7.6 & - & - & 34.9 & 75.0 & - & - & - & - 
    \\
    $\bullet$~FlowNav~\cite{gode2024flownav} & - & - & 36.0 & 64.1 & 9.2 & 8.5 & - & - & 35.8 & 75.7 & - & - & - & - 
    \\
    $\bullet$~NavDP~\cite{cai2025navdp}  & 77.8 & 74.8 & 72.5 & 167.2 & 43.4 & 43.4 & 38.7 & 31.9 & 43.6 & 93.7 & 59.3 & 21 & 5.5 & 3.5 
    \\
    \midrule
    \nickname\textbf{(Ours)} & \textbf{80.9} & \textbf{78.6} & \textbf{73.7} & \textbf{172.8} & \textbf{44.0} & \textbf{44.4} & \textbf{42.4} & \textbf{20.6} & \textbf{55.1} & \textbf{128.7} & \textbf{64.0} & \textbf{16} & \textbf{4.4} & \textbf{4.0}
    \\
    \bottomrule
\end{tabular}
}
\vspace{-0.4cm}
\end{table*}

\subsection{Experiment Setup}
\noindent\textbf{Baseline Models.}
We conduct comparisons with representative baselines across five distinct paradigms.
\textit{(1) Traditional Planning}: DWA~\cite{fox2002dynamic} and FBE~\cite{yamauchi1997frontier} are adopted as model-based, non-learned methods for local motion planning and frontier exploration.
\textit{(2) Reinforcement Learning}: DDPPO~\cite{wijmans2019dd} and Falcon~\cite{gong2025cognition} represent end-to-end reinforcement learning policies for point navigation and social navigation tasks.
\textit{(3) Hybrid Learning-Planning}: iPlanner~\cite{yang2023iplanner} and ViPlanner~\cite{roth2024viplanner} combine learned visual perception with classical planning pipelines.
\textit{(4) Imitation Learning}: GNM~\cite{shah2022gnm} and ViNT~\cite{shah2023vint} leverage large-scale visual pre-training to learn goal-conditioned navigation policies.
\textit{(5) Generative Modeling}: NoMaD~\cite{sridhar2024nomad} and NavDP~\cite{cai2025navdp} employ diffusion models for decision-making in navigation, while FlowNav~\cite{gode2024flownav} introduces flow matching into the navigation domain.

\noindent\textbf{Implementation Details.} 
We utilize a 16-layer Transformer decoder initialized with pre-trained weights from NavDP~\cite{cai2025navdp}, taking 8-frame RGB history and current depth as observations. Training follows a two-stage paradigm. In the first stage, we fine-tune the backbone on a 3,000-trajectory subset of Gibson scenes~\cite{xia2018gibson}---representing only $\sim$5\% of the data used by NavDP (nearly 56k trajectories)---using AdamW with learning rate $lr=5 \times 10^{-5}$ for 30 epochs. In the second stage, we apply GRPO to further refine the policy: 16 trajectories are sampled per update across a mixture of dynamic tasks, with episode-level normalized advantages. The policy is optimized without an explicit critic network ($lr=5 \times 10^{-6}$), with gradient updates triggered at episode boundaries. All experiments are conducted on a single NVIDIA RTX 4090 GPU.

\subsection{Main Results}
\label{sec:main_exp}

Table~\ref{tab:nav_comparison} summarizes our main results across six navigation tasks. \nickname achieves SOTA performance in both static and dynamic regimes across all reported metrics.

\noindent\textbf{Static Scenes.} Our method consistently outperforms all baselines across \textit{PointNav}, \textit{Exploration}, and \textit{ImageNav} tasks. Compared to previous SOTA baseline NavDP \cite{cai2025navdp}, we improve PointNav SR/SPL by \textbf{3.1\%/3.8\%}, Exploration ET/EA by \textbf{1.2\,m/5.6\,m\textsuperscript{2}}, and ImageNav SR/SPL by \textbf{0.6\%/1.0\%}. Notably, our approach outperforms traditional planning and learning-based methods by over 10\% in SR. 

\noindent\textbf{Dynamic Scenes.} All methods suffer performance decay when moving to dynamic scenes, yet \nickname demonstrates superior robustness. Notably, our performance gains over baselines are even more significant in dynamic environments than in static ones. For \textit{Dyn. PointNav}, SR improves \textbf{from 38.7\% to 42.4\%} while S-TL drops substantially from 31.9 to 20.6, indicating that our policy generates more effective paths without sacrificing success rate. The most notable gains appear in \textit{Dyn. Exploration}, where ET and EA improve by 26.4\% and 37.4\% respectively. For \textit{SocialNav}, SR increases \textbf{from 59.3\% to 64.0\%}, with Coll. reduced \textbf{from 21\% to 16\%} and SC improved from \textbf{5.5\% to 4.4\%}, demonstrating more socially-aware behavior learned from dynamic scenes.

Overall, our method achieves consistent superiority across all six tasks within a single unified policy, eliminating the need for task-specific reward engineering or hand-crafted cost functions. We observe that the performance gap over diffusion-based baselines expands in dynamic settings. This suggests that the straight-line probability flow of \nickname is particularly advantageous for real-time navigation: by linearizing the coupling between noise and action, our model provides more deterministic and responsive control signals, which are critical for maintaining safety and efficiency.


\begin{figure*}[t]
    \centering
    \includegraphics[width=\linewidth]{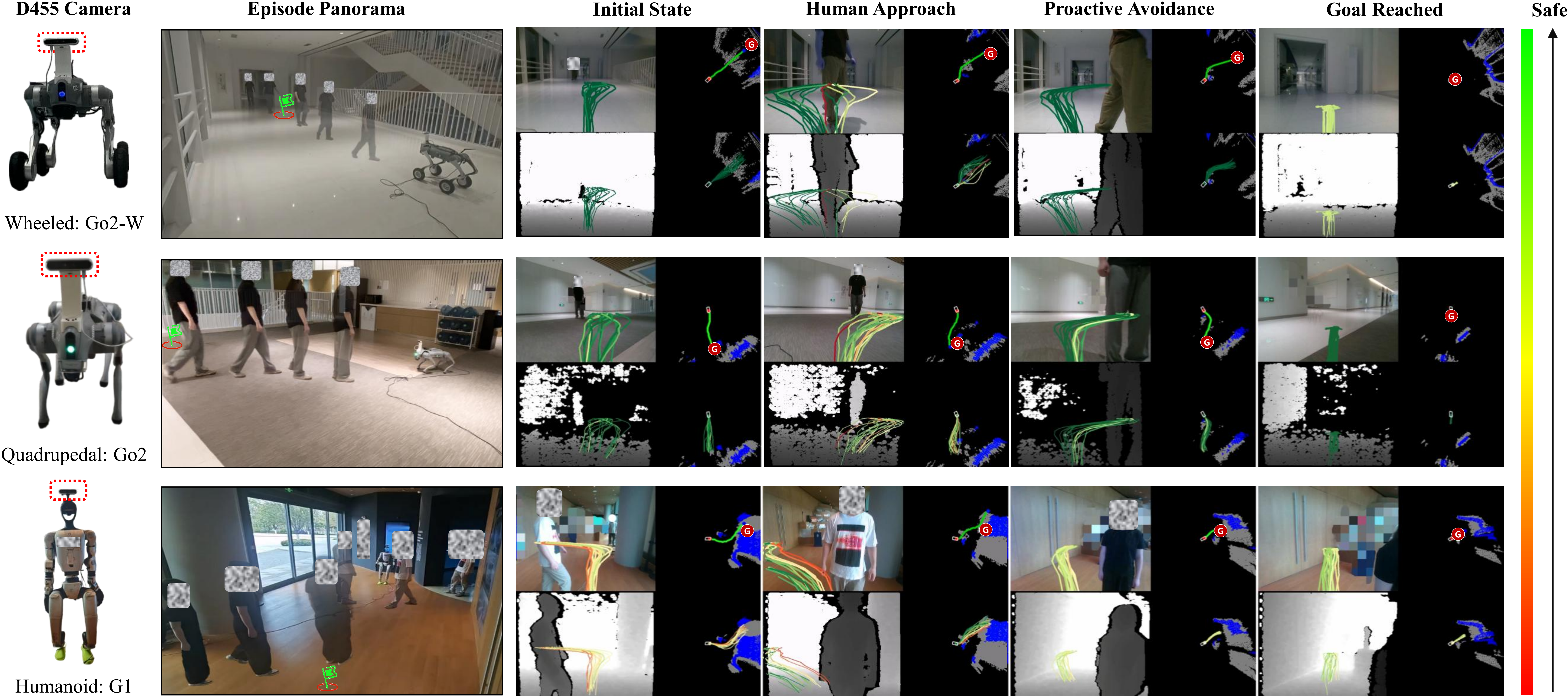}
    \vspace{-0.4cm}
    \caption{\textbf{Real-World Cross-Embodiment Deployment.} \nickname is deployed on three heterogeneous platforms—wheeled (Go2-W), quadrupedal (Go2), and humanoid (G1)—using only a single RGB-D camera. Despite significant differences in morphology and locomotion, our unified policy achieves robust zero-shot transfer without platform-specific fine-tuning. Critic-guided visualizations further show that safer paths with larger clearance from obstacles receive higher scores, demonstrating reliable risk-aware navigation in real-world scenarios.}
    \label{fig:deployment}
    \vspace{-0.2cm}
\end{figure*}

\begin{table*}[ht]
\caption{\textbf{Ablation studies} on \nickname. We report metrics for tasks most relevant to each component: \textit{PointNav} and \textit{Exploration} evaluate IL fine-tuning, while \textit{SocialNav} and \textit{Dyn. Exploration} assess RL post-training. $^\ddagger$ indicates excluded setups.}
\label{tab:ablation}
\centering
\resizebox{\linewidth}{!}{
\begin{tabular}{l|c|c|cc|cccc|cc|cc}
\toprule
\multirow{2}{*}{\textbf{Method}}  & \multirow{2}{*}{\textbf{Head Type}} & \multirow{2}{*}{\textbf{Training}}
& \multicolumn{2}{c|}{\textbf{PointNav}} & \multicolumn{4}{c|}{\textbf{SocialNav}} & \multicolumn{2}{c|}{\textbf{Exploration}} & \multicolumn{2}{c}{\textbf{Dyn. Exploration}} 
\\
& &
& SR$\uparrow$ & SPL$\uparrow$
& SR$\uparrow$ & Coll.$\downarrow$ & SC$\downarrow$ & MinDist.$\uparrow$
& ET$\uparrow$ & EA$\uparrow$
& ET$\uparrow$ & EA$\uparrow$
\\
\midrule
\midrule
\color{gray} $-$ Determinisitic Baseline$^\ddagger$ & \color{gray} Regression & \color{gray} IL & \color{gray} 53.6 & \color{gray} 50.0 & \color{gray} 49.3 & \color{gray} 18 & \color{gray} 4.9 & \color{gray} 3.8 & \color{gray} 32.8 & \color{gray} 48.1 & \color{gray} 30.5 & \color{gray} 71.0 
\\
DDPM Baseline  & Diffusion & IL &          
77.8 & 74.8 &
59.3 & 21 & 5.5 & 3.5 &
72.5 & 167.2 &
43.6 & 93.7 \\
$+$ Conditional Flow Matching   & Flow & IL&          
78.9& 75.6&
60.6 & 20 & 5.9 & 3.3 &
71.8& 166.8&
43.8 & 95.2 \\
$+$ Rectified Flow & Flow & IL &          
80.4 & 78.0 &
62.8 & 17 & 5.8 & 3.6 &
72.8 & 168.5 &
48.0 & 106.4 \\
$+$ GRPO Post-Training  & Flow & IL $+$ RL &          
80.9 & 78.6 &
64.0 & 16 & 4.4 & 4.0 &
73.7 & 172.8 &
55.1 & 128.7 \\
\bottomrule
\end{tabular}
}
\vspace{-0.4cm}
\end{table*}

\subsection{Real-World Deployment}
\label{sec:realworld}

To evaluate the cross-embodiment generalization of \nickname in the real world, we deploy our policy on three morphologically distinct platforms: a wheeled robot (Unitree Go2-W), a quadrupedal robot (Unitree Go2), and 
a humanoid robot (Unitree G1). These platforms pose fundamentally different deployment challenges. The wheeled platform offers stable, holonomic motion but lacks terrain adaptability. The quadrupedal introduces periodic gait 
cycles that cause camera oscillation and body sway, corrupting the visual input stream. The humanoid presents the greatest challenge: bipedal locomotion induces significant pitch and roll instability, and its elevated camera height (around 1.3\,m above ground) creates a substantially different visual perspective compared to the low-mounted cameras in the training data. All platforms are equipped with an Intel RealSense D455 RGB-D camera as the sole sensing modality, and inference runs on a remote machine with an RTX 4060 GPU. Experiments are conducted across diverse scenes including corridors, offices, and halls, in the presence of both static obstacles and dynamic pedestrians, as shown in Fig.~\ref{fig:deployment}.

Our deployment confirms that \nickname achieves effective sim-to-real transfer across wheeled, quadrupedal, and humanoid morphologies without any platform-specific fine-tuning. Furthermore, the critic-guided trajectory visualization reveals that our policy consistently assigns higher safety scores to paths that maintain greater clearance from pedestrians and obstacles, demonstrating reliable risk-aware trajectory selection in real-world dynamic environments.

\subsection{Ablation Studies}
\label{sec:abla_exp}

Tab.~\ref{tab:ablation} step-by-step ablates our two-stage design.

\noindent\textbf{Deterministic vs. Generative.} Replacing our generative backbone with a direct regression baseline leads to a significant performance drop. This underscores the necessity of modeling the action space's inherent stochasticity to handle complex navigation tasks.

\noindent\textbf{Stage 1 Backbone Selection.}
Replacing DDPM with CFM yields inconsistent results, with \textit{SocialNav} SC degrading from 5.5\% to 5.9\%. However, substituting CFM with RF resolves this instability, achieving consistent gains: \textit{PointNav} SR/SPL improves by 2.6\%/3.2\% and \textit{SocialNav} SR/Coll. by 3.5\%/4\%, validating RF's trajectory consistency.

\noindent\textbf{Stage 2: GRPO Refinement.}
RL post-training further boosts performance across all task regimes. In dynamic scenes, \textit{Dyn. Exploration} ET/EA rises from 48.0/106.4 to 55.1/128.7, while \textit{SocialNav} SR reaches 64.0\%. Crucially, as static environments represent a degenerate case of dynamic ones, the RL-refined policy also shows improved robustness in static tasks. This demonstrates that RL fine-tuning not only optimizes task-level objectives in crowded spaces but also strengthens the underlying geometric priors, leading to a more generalized navigation policy.

\subsection{Efficiency \& Sensitivity Analysis}
\label{sec:efficiency}

We evaluate the computational efficiency of \nickname against generative baselines in Tab.~\ref{tab:efficiency}. Under the same sampling budget ($K{=}10$), \nickname$^\dagger$ reduces inference latency by 19.8\% and 40.1\% compared to NavDP and FlowNav, respectively, attributed to rectified flow's linearized probability transport which reduces per-step computational overhead. Moreover, \nickname maintains high-quality trajectory generation with only $K{=}\mathbf{6}$ steps, achieving a 29.1\% speedup over NavDP, which is critical for rapid replanning in dynamic environments.


\begin{table}[t]
\centering
\caption{\textbf{Inference efficiency comparison of generative navigation policies.} \nickname$^\dagger$ uses 10 Euler steps for fair comparison.}
\label{tab:efficiency}
\resizebox{\linewidth}{!}{
\begin{tabular}{lccc}
\toprule
\textbf{Method} & \textbf{Type} & \textbf{Steps} $K$ & \textbf{Inference Time (ms)} \\
\midrule
NoMaD~\cite{sridhar2024nomad}     &  \multirow{2}{*}{\textbf{Diffusion-based}} & 10 & 468.4 \\
NavDP~\cite{cai2025navdp}             &  & 10 & 305.2 \\
\midrule
FlowNav~\cite{gode2024flownav}         & \multirow{3}{*}{\textbf{Flow-based}}      & 10 & 408.6 \\
\nickname$^\dagger$               &     & 10 & 244.9 \\
\nickname     &       &  \textbf{6} & \textbf{216.5} \\
\bottomrule
\end{tabular}
}
\vspace{-0.4cm}
\end{table}

\begin{figure}[t]
    \centering
    \includegraphics[width=\columnwidth]{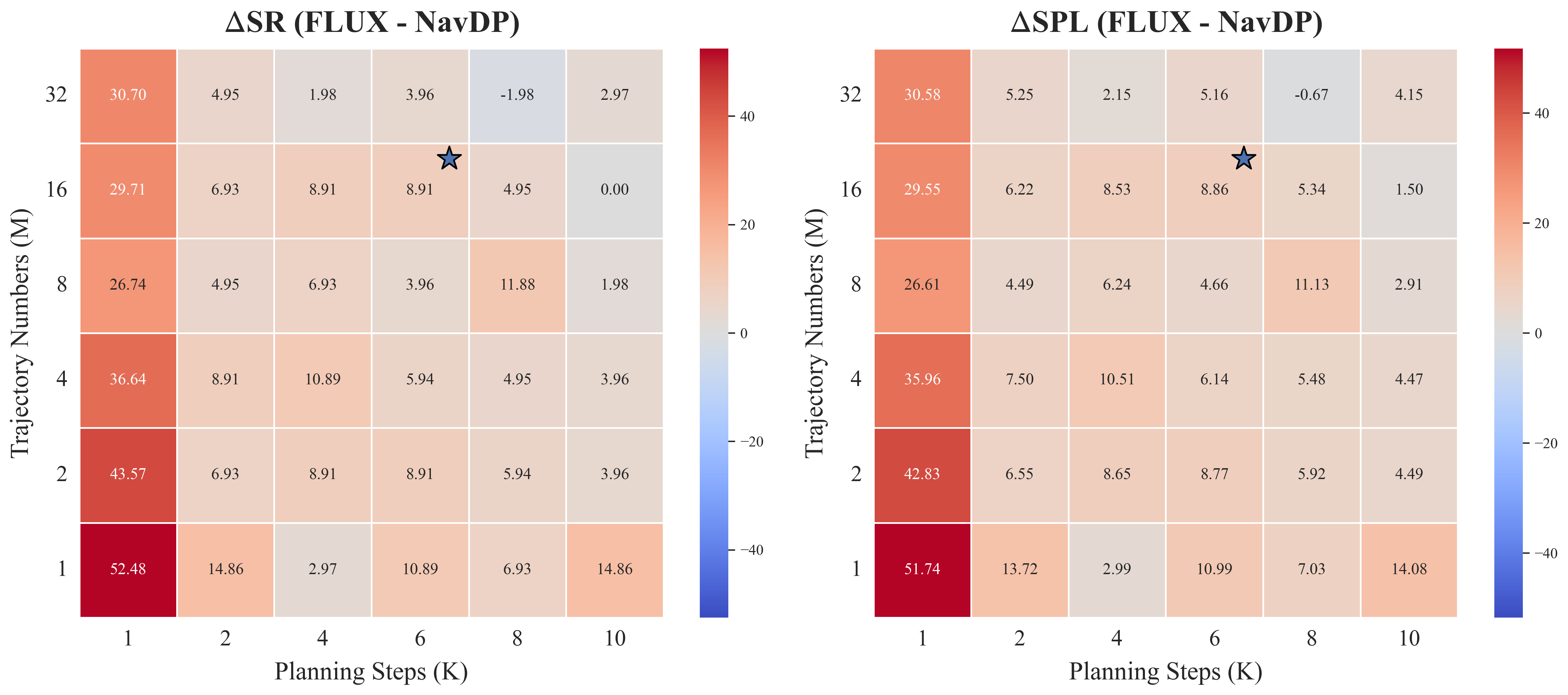}
    \vspace{-0.4cm}
    \caption{\textbf{Sensitivity analysis of trajectory number $M$ and sample steps $K$}. \nickname outperforms NavDP across nearly all $(M, K)$ configurations, with $M{=}16$, $K{=}6$ achieving near-peak performance.}
    \label{fig:sensitive}
    \vspace{-0.4cm}
\end{figure}

Fig.~\ref{fig:sensitive} analyzes the sensitivity of \nickname on \textit{PointNav}. \nickname{} outperforms NavDP across nearly all configurations, with only marginal degradation observed at the extreme setting of $M{=}32$, $K{=}8$. Performance improves as $M$ scales from 1 to 16 and saturates beyond, while larger $K$ yields diminishing returns at increased inference cost. We adopt $M{=}16$ and $K{=}6$ as default, with a tradeoff of near-peak performance and real-time efficiency.

\subsection{Qualitative Analysis} 
\label{sec:qualitative}

\begin{figure}[t]
    \centering
    \includegraphics[width=\linewidth]{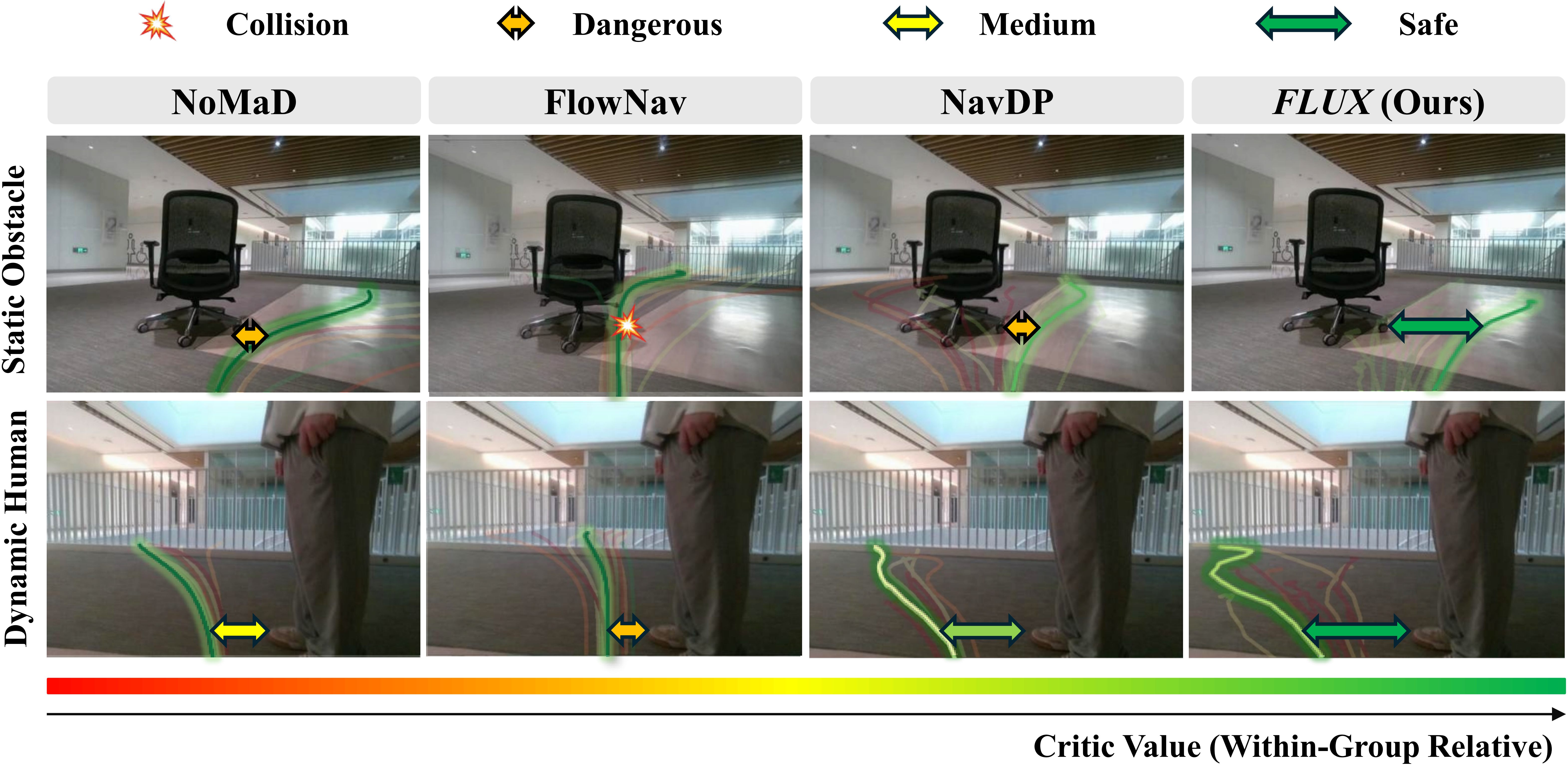}
    \vspace{-0.4cm}
    \caption{\textbf{Qualitative analysis of generative navigation policies.} Compared to prior methods, \nickname generates trajectories with higher safety in both static and dynamic obstacle avoidance.}
    \label{fig:qualitative}
    \vspace{-0.4cm}
\end{figure}

Fig.~\ref{fig:qualitative} compares trajectory generation across generative baselines. In static obstacle avoidance, FlowNav suffers from a collision, while NoMaD and NavDP generate trajectories with marginal safety clearances. In contrast, \nickname identifies and selects a high-clearance, risk-free path. For dynamic human interaction, whereas NoMaD and FlowNav exhibit medium-risky behaviors, both NavDP and \nickname perform proactive avoidance. Notably, \nickname demonstrates a more decisive lateral detour, reflecting superior social awareness and risk-sensitive planning compared to prior policies.

\section{Limitations}

Despite the strong performance of \nickname{}, several limitations remain. First, the policy lacks explicit social reasoning, such as interpreting non-verbal human intentions or adhering to implicit etiquette, which limits trajectory interpretability. Second, current evaluations are primarily constrained to pedestrian-filled indoor spaces using simulated agents; extending \datasetname{} to more diverse environments and heterogeneous dynamic agents remains an open challenge. Future work will focus on integrating real-world social interaction datasets to further bridge the sim-to-real gap.

\section{Conclusion}
We presented \nickname{}, a unified fundamental navigation framework that utilizes rectified flow to achieve efficient control in both simulation and real world. Unlike diffusion models, the linearized probability transport of \nickname{} enables faster convergence and superior reliability, which are critical for safe social navigation. Our results indicate that \nickname{} not only sets new SOTA records across diverse tasks in our proposed \datasetname{} benchmark but also exhibits cross-embodiment capability in the real world. 


\bibliographystyle{IEEEtran} 
\bibliography{IEEEexample}

\end{document}